\documentclass[conference]{IEEEtran}
\IEEEoverridecommandlockouts
\usepackage{cite}
\usepackage{amsmath,amssymb,amsfonts}
\usepackage{algorithmic}
\usepackage{graphicx}
\usepackage{textcomp}
\usepackage{xcolor}
\usepackage{url}
\def\BibTeX{{\rm B\kern-.05em{\sc i\kern-.025em b}\kern-.08em
    T\kern-.1667em\lower.7ex\hbox{E}\kern-.125emX}}
\begin{document}

\title{Baby Physical Safety Monitoring in Smart Home Using Action Recognition System \\
}

\author{\IEEEauthorblockN{Victor Adewopo}
	\IEEEauthorblockA{\textit{School of Information Technology} \\
		\textit{University of Cincinnati}\\
		Cincinnati, OH \\
		adewopva@mail.uc.edu}
	\and
	\IEEEauthorblockN{Nelly Elsayed}
	\IEEEauthorblockA{\textit{School of Information Technology} \\
		\textit{University of Cincinnati}\\
		Cincinnati, OH \\
		elsayeny@ucmail.uc.edu}
    \and
    \IEEEauthorblockN{Kelly Anderson}
    \IEEEauthorblockA{\textit{Corporate R\&D Data Science \& AI} \\
    \textit{Procter \& Gamble (P\&G)}\\
    Cincinnati, Ohio, United States \\
    anderson.kl.1@pg.com}}


\thispagestyle{empty}

\begin{huge}
IEEE Copyright Notice
\end{huge}

\vspace{5mm} 

\begin{large}
Copyright (c) 2022 IEEE
\end{large}

\vspace{5mm} 

\begin{large}
Personal use of this material is permitted. Permission from IEEE must be obtained for all other uses, in any current or future media, including reprinting/republishing this material for advertising or promotional purposes, creating new collective works, for resale or redistribution to servers or lists, or reuse of any copyrighted component of this work in other works.
\end{large}

\vspace{5mm} 

\begin{large}
\textbf{Accepted to be published in:} IEEE SoutheastCon 2023; April 13 - 16, 2023 - Orlando, Florida, USA.
https://ieeesoutheastcon.org/

\end{large}

\vspace{5mm} 



\maketitle

\begin{abstract}
Humans are able to intuitively deduce actions that took place between two states in observations via deductive reasoning. This is because the brain operates on a bidirectional communication model, which has radically improved the accuracy of recognition and prediction based on features connected to previous experiences. During the past decade, deep learning models for action recognition have significantly improved. However, deep neural networks struggle with these tasks on a smaller dataset for specific Action Recognition (AR) tasks. As with most action recognition tasks, the ambiguity of accurately describing activities in spatial-temporal data is a drawback that can be overcome by curating suitable datasets, including careful annotations and preprocessing of video data for analyzing various recognition tasks. In this study, we present a novel lightweight framework combining transfer learning techniques with a Conv2D LSTM layer to extract features from the pre-trained I3D model on the Kinetics dataset for a new AR task (Smart Baby Care) that requires a smaller dataset and less computational resources. Furthermore, we developed a benchmark dataset and an automated model that uses LSTM convolution with I3D (ConvLSTM-I3D) for recognizing and predicting baby activities in a smart baby room. Finally, we implemented video augmentation to improve model performance on the smart baby care task. Compared to other benchmark models, our experimental framework achieved better performance with less computational resources.
\end{abstract}


\section{Introduction}
Action Recognition (AR) is a revolutionary topic in machine learning and computer vision that has been utilized for advance high-precision applications such as video surveillance, medical imaging, and digital signal processing.  
Training Action Recognition (AR) models require a massive dataset and high computational resources. Kunze et al.~\cite{Kunze2017} proved that inferring knowledge from existing models saves time taken in training new models and reduction in cost while maintaining a significant accuracy with constrained GPU.
Humans can recognize activity around them based on a set of repeated features that have been 
registered in the subconscious mind. Analyzing motion is a very common topic in computer vision which involves the movement of objects and their behavioral change, which varies depending on a specific area of application. Typical areas of interest include tracking objects, deformation quantification, and detecting abnormal behavioral patterns~\cite{Fortun2015}. Action Recognition also entails localizing objects where the action of interest is performed. This is a big challenge in computer vision because it requires the action class to be correctly identified by determining its spatiotemporal location. Action Localization and Action Recognition in videos are gray areas and do not necessarily mean the same thing. Action localization can be more challenging because the action class must be correctly identified and determine its spatiotemporal location. Only 15\% of research papers on benchmark datasets discussed action localization results~\cite{Li2020}.
Action Recognition in videos has two crucial key points; Appearances and Temporal dynamics. ConvNets was introduced to solve the challenge of video-based action recognition. However, this was affected by long-range temporal structures focusing on appearances and shot term motions up to 16 frames. Previous models were built on perfectly trimmed video, which is sharply in contrast with untrimmed video in real-life scenarios. The lack of a large volume training dataset and dominating background may affect the performance of Action Recognition tasks. The research of Wang et al.~\cite{Wang2017}, proposed a temporal segment network (TSN), a flexible video-level framework for learning action models in the video.

Developing an Action Recognition model in a different application domain such as the medical field by interpreting video contents can improve precision in of Human Assisted AI in the medical context, i.e., deformation of cells can easily be spotted and analyzed using AI tailored toward this specific use case by estimating and training models on larger samples~\cite{Fortun2015}. Most surgeries can be performed by human-assisted AI, thereby eliminating medical errors and ensuring precision in medical procedures and diagnosis. Center for Disease Control and Prevention reported a decline in Sudden Unexpected Infant Death and Sudden Infant Death Syndrome (SUIDS/SIDS), sometimes known as crib death, from 154 deaths per 100,000 live births in 1990 to 90 deaths per 100,000 live births in 2019. Today, approximately 3,400 babies in the United States die of SUIDS each year, with 28\% of the deaths caused by accidental suffocation and strangulation in bed~\cite{CDC2021}. The main contribution of this paper is to develop a specific Action Recognition model for smart baby care and monitoring of babies' activities in the crib while providing real-time updates to parents and caregivers' handheld devices on current and future temporal actions in the crib, which is a new application area.
We curated a novel dataset from scratch, which was implemented with our experimental framework. Our framework proved that specific Action Recognition tasks could be developed with a relatively small dataset and reduced computational resources on a constrained GPU by employing data augmentation and transfer learning methodologies. The contributions of this work can be summarized as follows:

\begin{enumerate}
	\item An IoT system for monitoring the baby physical safety via an online smart surveillance camera data stream.
	\item An automated deep learning model based on convolution LSTM and I3D transfer learning (ConvLSTM-I3D) for tracking and monitoring baby activities that empirically has a comparable performance to the state-of-the-art action recognition models.
	\item A unique smart baby room dataset that contains five different action classes of baby activities in the crib that could be utilized in several computer vision-based tasks.
\end{enumerate}

The structure of this paper is organized as follows: Section~\ref{sec:HumanAction} describes the human action recognition task challenges.Section~\ref{sec:transferLearning} provides a brief description of transfer learning methodologies, focusing on specific action recognition tasks. Section~\ref{sec:ModelArchitecture} describes the proposed model architecture. Section~\ref{sec:resuts} provides the results and analysis.

\section{Human Action Recognition} \label{sec:HumanAction}
Deep learning algorithms achieved high-performance results in recognizing and classifying actions in the short trimmed video. The research of Li et al.~\cite{Li2020} focused on segmenting and classifying activities in long untrimmed videos using multi-stage architecture for the temporal action segmentation task with a dual dilated layer that combines both large and small receptive fields in capturing long-range dependencies and recognizing action segments.
The multi-stage architecture takes into consideration predictions from the previous stage as input, each stage in the network makes a prediction by refining the output of the last stage in a frame-wise manner represented as; $(Y^0 = x_{1:T})$ and $(Y^s = F(Y^{{s}-{1}}))$, where, $(Y^s)$ is the output at stage $(S)$, $(F)$ stands for the single-stage model with only temporal convolution layers~\cite{Li2020}.
Stacking models (multi-stage architectures) sequentially increased the performance of computer vision tasks as each model operated directly on the output of the previous models. The stacked hourglass networks for human pose estimation utilized bottom-up processing, a high to low-resolution top-down processing low to high resolution in their network architecture~\cite{Newell}.

\subsection{Challenges of Action Recognition Task}
Inferring knowledge automatically from actions in the video stream is an essential futuristic direction in computer vision tasks. Developing AR models for untrimmed videos and accurately recognizing the action performed is more beneficial in real-life scenarios than classifying action from a short trimmed video. 
Action localization is generally a more difficult task as it requires the action class to be correctly recognized and also its spatiotemporal location to be identified. The experimental setup in~\cite{Soomro2014} on a realistic sports dataset (UCF sports dataset)~\cite{Soomro2014} containing ten actions and 150 clips for action localization used the Leave One Out (LOO) scheme, which trains the dataset on nine actions and leaves only one out. This approach was criticized with experimental backup that video may have a similar background and consequentially yield higher accuracy due to high correlation between videos\cite{lan2011discriminative}. Extracting features such as data points that carry distinguishing information about actions being performed becomes essential and can be achieved in two ways:
\begin{enumerate}
	\item \textit{Low-level extraction:} This detects related interest points using sparse key points such as corners, edges, blobs, and so forth. Space-Time Interest Points (STIP) were developed to capture features in the video.
	\item \textit{High-level extraction:} This captures more specific information such as pose, shape, and contextual information. 
\end{enumerate}
The action localization task can be achieved by using the sliding window approach. This method slides in the spatiotemporal volume of the video finds the subvolume with the highest scores, and assigns a bounding box.
Most action recognition research work utilized three-step in video classification tasks:
\begin{enumerate}
	\item Features extraction,
	\item Dictionary learning, and 
	\item Classification.
\end{enumerate}
Temporal relational reasoning in video streams focuses on inferring the relationship between transformation/temporal dependencies between two video frames. Zhou et al.~\cite{Zhou} proposed Temporal Relation Network to enable possible temporal relations between observations on benchmark datasets (Something-Something, Jester, and Charades).
Deep CNN has helped significantly in improving the performance of image classification tasks, and recognition of action in video stream remains a core challenge of computer vision. 
The 3D CNN has limited sequences of frames fed into the computationally expensive network; the Temporal Relation Network address this challenge and improves model performance by a wide margin by sampling individual frames to understand the causal relationship in contrast to the research work in \cite{pirsiavash2014parsing,gaidon2014activity} which modeled temporal structures for action recognition using bag of words, motion atoms, and action grammar.

\section{Methods of Transfer Learning} \label{sec:transferLearning}
Temporal relational reasoning in video streams focuses on inferring the relationship between transformation/temporal dependencies between two video frames. Zhou et al.~\cite{zhou2018temporal} proposed Temporal Relation Network to enable possible temporal relations between observations on benchmark datasets (Something-Something, Jester, and Charades).
Deep CNN has helped significantly improve the performance of image classification tasks, and recognition of action in video stream remains a core challenge of computer vision. 
The 3D CNN has limited sequences of frames fed into the computationally expensive network; the Temporal Relation Network address this challenge and improves model performance by sampling individual frames to understand the causal relationship in contrast to the research work in \cite{pirsiavash2014parsing,gaidon2014activity} that modeled temporal structures for action recognition using a bag of words, motion atoms, and action grammar.

The \textit{Heterogeneous Transfer Learning} trains a base model on multiple datasets with different tasks simultaneously where the domains have different feature spaces from the target dataset~\cite{wang2015transfer}. Chen et al.~\cite{chen2015} proposed a multitask learning approach to improve low-resource automatic speech recognition using a deep neural network without requiring additional language resources. Their experimental model achieved a very competitive result. Training model using Heterogeneous transfer learning approach has the potential to solve data issues and close the performance gap in the new target domain with different source feature spaces. However, Heigold et al.~\cite{Heigold} research on multilingual acoustic models using distributed DNN showcased that there is still need for large amount of data to achieve significant improvements in model performance.
Heterogeneous transfer learning is a relatively new area of study. The primary area of application revolves around image classification,multi-language text classification, drug efficacy classification, and human activity classification. The two main approaches used in solving heterogeneous feature space differences are symmetric and asymmetric feature-based transfer learning~\cite{weiss2016survey}. The research of Harel et al.~\cite{harel2010learning} proposed a framework for learning 
 single task with data provided from different feature spaces called an ``outlook". A customized wearable sensor system used a heterogeneous transfer learning approach for human activity classification. Their result indicated performance boost in mapped data compared to partial data in the target outlook.

The \textit{Domain Adaptation Transfer Learning} trains the model on one or more tasks afterward retrains the whole model layers or some part of the layers on a new task different from the source domain. The process adapts one or more source domains to transfer knowledge and improve the performance of the target
domain. The features learned in the base model serve as the starting point for training on a new task. Our experimental research utilized the domain adaptation transfer learning approach by freezing part of the source domain (base model layer) and focused on features similar to our target domain (specific task) to avoid overfitting our model. The domain adaptation process attempts to alter a source domain to bring the distribution of the source closer to that of the target~\cite{weiss2016survey}. The research of Hal Daum{\'e}~\cite{daume2009frustratingly} utilized domain adaptation approach on named entity recognition and outperforms the state-of-the-art on a range of datasets. Domain adaptation can be either supervised or unsupervised learning. The main difference between supervised and unsupervised learning is that the dataset in the supervised learning target domain is labeled, while in unsupervised learning, the dataset is not labeled; the source domain model generally labels the data before domain adaptation. Li et al.~\cite{Li2020} fine-tuned pre-trained deep neural networks on image classification and video analysis tasks. Lim et al.~\cite{Lim2018} worked on fine-tuning and feature extraction of pre-trained models for Facial expression recognition. VGG-Net and ResNet were used as the source domain. Training most deep neural network models requires high computational costs and expensive GPU training models. Transfer learning on the twelve specific action classes considered in this research exploited the learned features from the source domain and produced better results with an \(F_1\) score of 74\%.

\subsection{Transfer learning for Specific Action Recognition Tasks}
The research of Kunze et al.~\cite{Kunze2017} applied transfer learning techniques and lower resource footprint to adopt an end-to-end CNN-based model trained on the English Language to recognize German speech.
Labeled data for a specific Action Recognition task with an adequate volume of data is empirically tricky. Specific AR tasks can benefit from the concept of transfer learning which involves enhancing the performance of an existing model by extracting features learned from the previous domain and applying them to a specific AR task with limited data availability and computing resources.
Ji et al.~\cite{Ji2013} proposed extracting temporal and spatial information from multiple adjacent frames through 3D convolutions. Human action recognition, specifically in the domain of behavior analysis and customer attributes, is a challenging task based on occlusion, camera angle, and clutter backgrounds. Visual object recognition has higher performance than the conventional pattern recognition paradigm with handcrafted features using appropriate CNN regularization. It was also observed that the performance differences between 3D CNN and other methods tend to be larger when the number of positive training samples is small.
In video analysis, spatial information is captured in the convolution stage to capture temporal features and spatial dimensions by convolving a 3D kernel to the cube formed by stacking multiple contiguous frames together, thereby capturing spatial information through connected multiple frames from previous convolution layers. Experiments were performed on three action classes cell to ear, object put, and pointing with the precision 64\%,67\%, and 82\%, respectively.

Another area of application for specific AR tasks is smart city surveillance, where violence can easily be spotted to alert appropriate enforcement agencies with automated analysis of video contents in surveillance cameras \cite{Fortun2015}.
Robotica and auto navigation has also benefited from the use of AR system for automatic guidance, specifically in obstacle detection, accident prevention, and lane departure assist \cite{Fortun2015,adewopo2022review}.

\begin{figure} 
	\centering
	\includegraphics[width=8cm, height = 4.5cm]{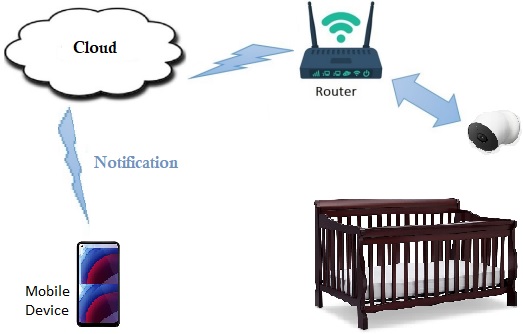}
	\caption{The proposed system architecture design for baby safety at home.}
	\label{fig:overall}
\end{figure}


\section{System Architecture Design}\label{sec:systemDesign}
The proposed baby safety system design is shown in Figure~\ref{fig:overall}. First, a smart surveillance camera will be set above the baby crib. Then the camera will send real-time obtained video to the cloud for the action recognition process. Due to the 5G technology, the speed of transferring the video will be significantly higher than the older technologies~\cite{gohil20135g,haring2021review}. The baby action recognition system will process the received video from the smart surveillance camera. The action recognition model will be connected to a smartphone device application connected to the Internet. Once there is an unsafe baby action (e.g., jumping from the crib), the smartphone device application will provide notification alert to the cellphone to alert guardian/parent of the child immediately to provide the child with prompt attention and prevent child injury or death. The alarm notification will turn the screen color red, high-tone ringing, and phone vibration. During the safe baby action monitoring, the application will provide the guardian/parent with a monitoring actions list in addition to real-time baby watching via surveillance camera.
\begin{figure} 
	\centering
	\includegraphics[width=8cm, height = 4.5cm]{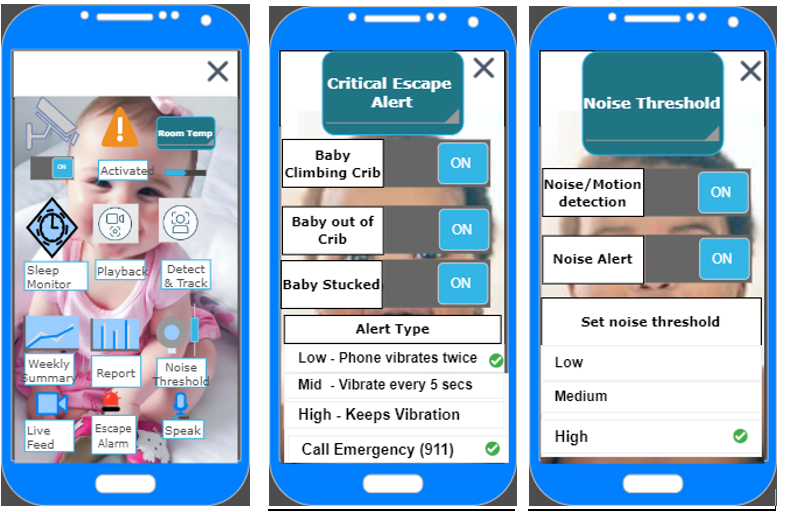}
	\caption{The proposed application interface design for smart baby monitor.}
	\label{mobileapp}
\end{figure}

In order to ensure infant safety and well-being, the developed model is integrated with a mobile application that can remotely monitor infants in the crib while caregivers perform other duties. A parent or caregiver can monitor their baby autonomously through the application and set customizable notification threshold. Figure~\ref{mobileapp} shows that the application architecture includes aesthetics that make it possible for users to switch between different modes for monitoring actions, including playback, live feed, and tracking, to monitor various activities that happen in the room autonomously without constantly looking at the handheld device.

\begin{figure*} 
	\centering
	\includegraphics[width=0.8\linewidth]{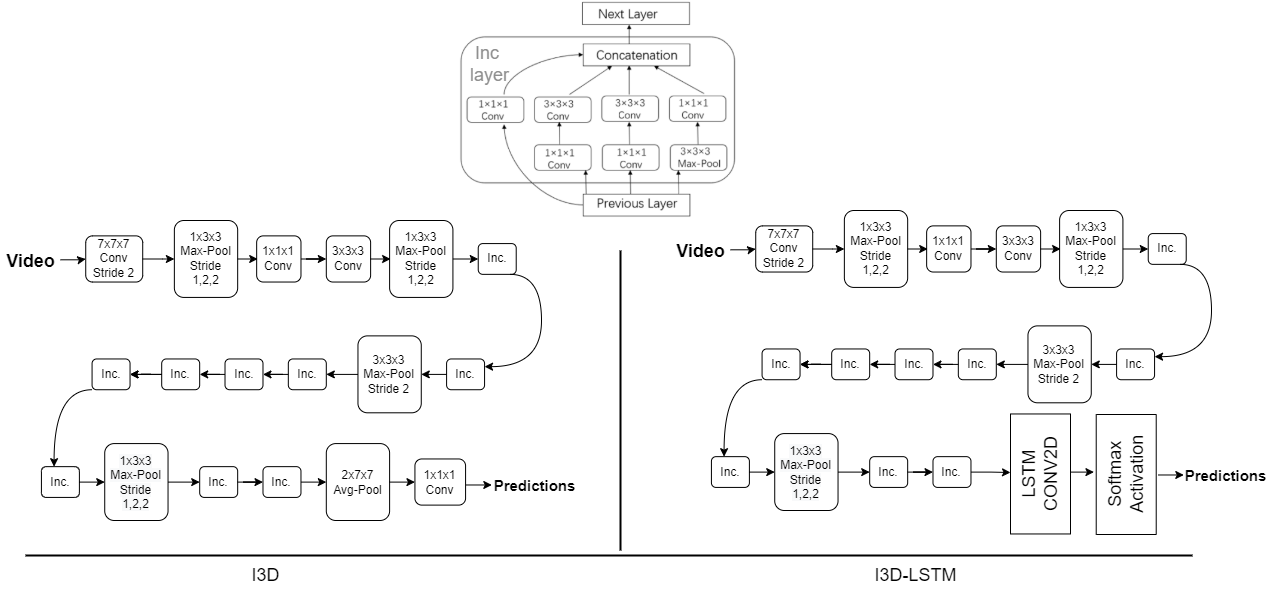}
	\caption{Model architecture for the proposed specific action recognition task.}
	\label{fig:f6}
\end{figure*}


\begin{figure} 
	\centering
	\includegraphics[width=8cm, height = 4.5cm]{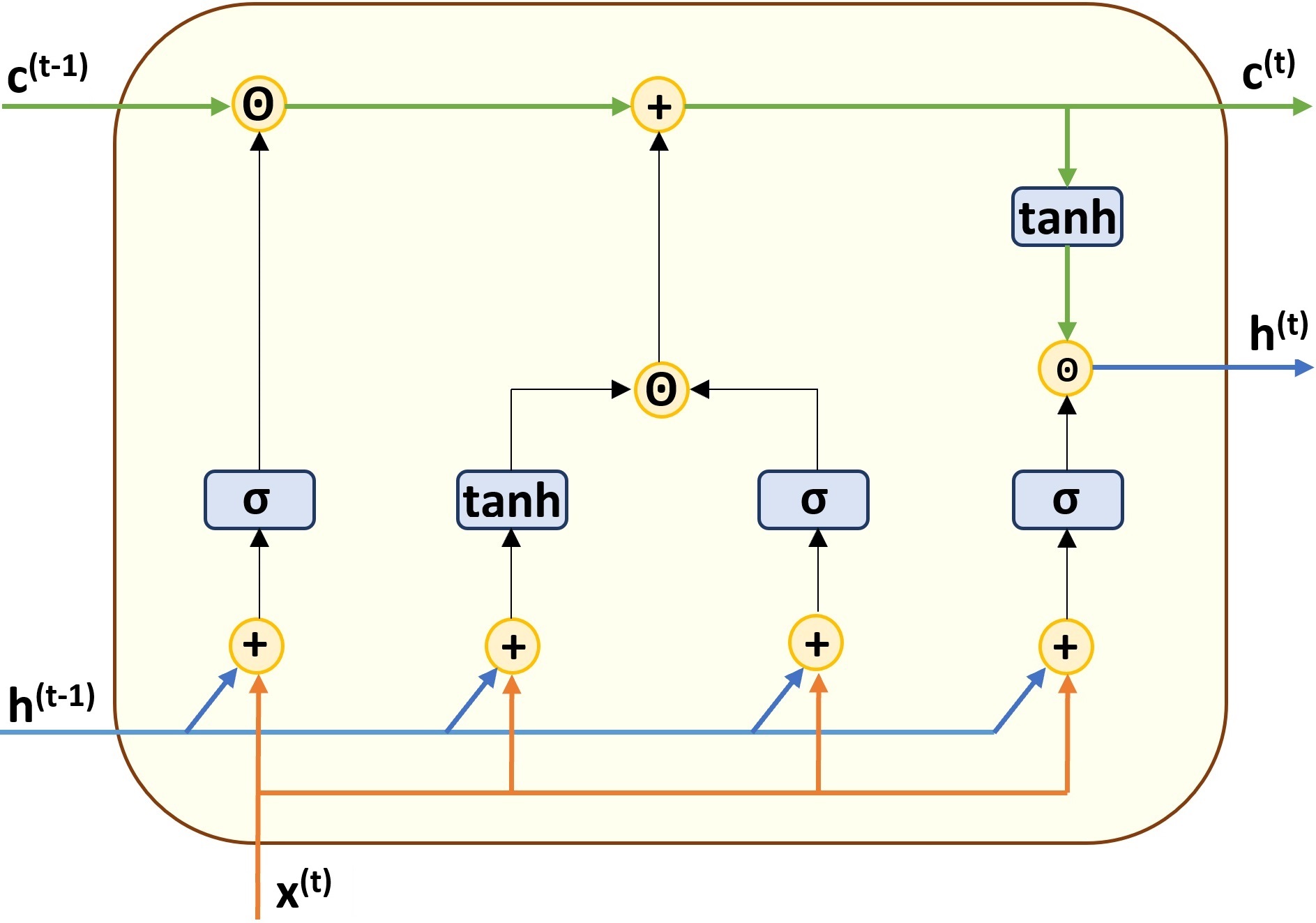}
	\caption{The convolution LSTM architecture~\cite{elsayed2020reduced}.}
	\label{fig:convLSTM}
\end{figure}

\section{Model Architecture}\label{sec:ModelArchitecture}

In our proposed model, the I3D model has been selected as the transfer learning portion of the model due to the performance of the I3D model on the benchmark dataset Kinetics 700 and the computational cost of the model, including hardware requirements. Other research has developed convolution-based architectures that achieved high performance in tasks such as image classification \cite{lim2016speech,adewopo2020exploring}. The shortcoming of CNN is the inability of the layers to extract temporal features; hence they are not efficient in handling complex tasks such as baby monitoring in a crib which requires extracting both temporal and spatial features \cite{elsayed2018deep,2021exploring}. Therefore, we combined the convolutional LSTM (ConvLSTM2D) layers with the I3D model. 
ConvLSTM2D architecture layers share the cell operations like the LSTM layers, except that the input of ConvLSTM2D architecture are convolutions compared to arithmetic operations in LSTM, and they have been proven to show higher performance with video compression and preprocessing~\cite{Robles-Serrano2021}. 

The ConvLSTM2D architecture layers share the cell operations similar to the LSTM layers, except that the internal arithmetic operations of the ConvLSTM2D are performed using the convolution multiplication concept and that the weights are replaced by the convolution kernels (filters) that are compatible with the convolution operations. Figure~\ref{fig:convLSTM} shows the ConvLSTM2D architecture~\cite{elsayed2020reduced,shi2015convolutional}. 
The output of the ConvLSTM2D $h^{(t)}$ at time $t$ is calculated by:
\begin{align}
	i^{(t)}&= \sigma(W_{xi} * x^{(t)} +U_{hi} * h^{(t-1)}+ W_{ci}\odot c^{(t-1)} + b_i)\label{eqn:i_convLSTM}\\ 
	g^{(t)}&= \mathrm{tanh}(W_{xg} * x^{(t)} +U_{hg} * h^{(t-1)}+ b_g)\label{eqn:g_convLSTM}\\ 
	f^{(t)}&=\sigma(W_{xf} * x^{(t)} +U_{hf} * h^{(t-1)} + W_{cf} \odot c^{(t-1)} + b_f)\label{eqn:f_convLSTM}\\ 
	o^{(t)}&=\sigma(W_{xo} * x^{(t)} +U_{ho} * h^{(t-1)} + W_{co} \odot c^{(t-1)} + b_o)\label{eqn:o_convLSTM}\\ 
	c^{(t)}&= f^{(t)}\odot c^{(t-1)} + i^{(t)} \odot g^{(t)}\label{eqn:s_convLSTM}\\
	h^{(t)}&= \mathrm{tanh}(c^{(t)})\odot o^{(t)}\label{eqn:h_convLSTM}
\end{align}
\noindent
where $i$, $f$, and $o$ are the input, forget, and output gates, respectively. $g^{(t)}$ is the input update. The symbol $*$ denotes the convolutional multiplications. The $W$s represent the feedforward convolutional kernel weights, and the $U$s represent the recurrent kernel weights. The biases are represented by $b$s. $c^{(t)}$ is the memory cell. 

The ConvLSTM2D has shown significant performance in various video processing due to its capability to learn spatial and temporal information from the video data.

The overall architecture of the I3D model is showcased in the original paper~\cite{Carreira2017} and our proposed modified framework is shown in Figure~\ref{fig:f6}.


	
	%


\begin{table*}
	\caption{The results of different action recognition models for baby action recognition.}
	
	\begin{center}
		\resizebox{\textwidth}{!}{
			\begin{tabular}{|l|c|c|c|c|c|c|c|c|}
				\hline
				\textbf{Model}&\textbf{\#Trainable}& \textbf{Train}&\textbf{Train}&\textbf{Cross}& \textbf{Test}& \textbf{Precision}&\textbf{Recall}&\textbf{F1-Score}\\
				\textbf{}&\textbf{\#Param}& \textbf{Time}&\textbf{Accuracy}&\textbf{Validation}& \textbf{Accuracy}& \textbf{}&\textbf{}&\textbf{}\\
				\hline
				I3D  & 9,440,261 & 0D 11H &\textbf{99\% }& \textbf{99\%}& \textbf{90\%}& 85\%& 85\%& 85\%\\
				\hline
				I3D-ConvLSTM2D  & 9,506,309 & \textbf{0D 10H} & \textbf{99\%} & \textbf{99\% }& \textbf{90\%}& \textbf{86\%}& \textbf{87\%}& \textbf{86\%}\\
				\hline
				I3D Video Aug.  & \textbf{2,360,069} & 1D 17H & 97\% & \textbf{99\%} & 84\%& 77\%& 75\%& 75\%\\
				\hline
				I3D-ConvLSTM2D-  &  9,472,005 &1D 10H& 93\% & 92\%& 85\%&80\%&81\%& 80\%\\
				Video-Aug.  &  && & & &&& \\
				\hline
		\end{tabular}}
		\label{Results}
	\end{center}
	
\end{table*}

\section{Dataset Collection}\label{AA}
Based on the uniqueness of our use case, we curated a dataset from scratch for this project to develop an AR model for specific Action Recognition task (Smart Baby Care). Our data source came from open-source video from social media platforms such as (YouTube, Instagram, Pexels, etc.). The videos were manually downloaded. We automate the trimming of videos with a python script using the inbuilt FF-MPEG library and annotated frame-wise using Labelbox tool \cite{labelbox}.
We developed a playlist for automatically generating links to video that matches our specified requirements. These videos were automatically downloaded using Python script and manually sorted to ensure the video is suitable for our use case.

\subsection{Features Extraction}
Extraction of features and dimensionality reduction are essential in increasing model performance and reducing computation cost~\cite{zebari2020comprehensive}. 
Feature extraction entails defining a set of features characteristics from an existing model that similar and meaningfully represent features in a new target domain \cite{ABHANG201697}.
A pre-trained model can be used standalone to extract features from a new dataset, and the extracted features can then be used as input in the development of a new model~\cite{Orenstein}.
We performed experiments with our dataset by performing feature extractions with the pre-trained I3D model before constructing a new model with the extracted features.
\subsection{Dataset Augmentation}
Neural Networks rely solely on big data to achieve high performance. Data augmentation is one of the techniques utilized in compensating for limited data in training a deep neural network. Data augmentations include flipping, rotating, cropping, adding noise, occluding portions of the image, and other techniques to enhance the size and quality of training datasets \cite{shorten2019survey}. 
Most of the data augmentation library focused on manipulating images. 
We performed framewise manipulation of each sample in our training dataset to have a completely transformed video stream by utilizing VidAug, a python library, to resize randomly, Horizontal flip, Rotate, Gaussian Blur, and Elastic transformation in our training dataset.  
The overarching goal of this approach is to develop a model that can perform well by augmenting data to overcome issues of camera angle, lighting, occlusion, background, scale, video orientation, and other invariances.
We developed a script to modify VidAug open-source code~\cite{Kopuklu}. Our developed script takes each
original video, reads its frames, and then applies the techniques discussed above with VidAug. Performing video augmentation is
more precarious as it involves applying exact changes to all frames in a single video sample, as discussed in~\cite{domenech2020sing}. The shortcomings of this approach are a 30\% increase in computation cost and time.



\begin{figure*} [!htbp]
	\centering
	\includegraphics[width=\linewidth]{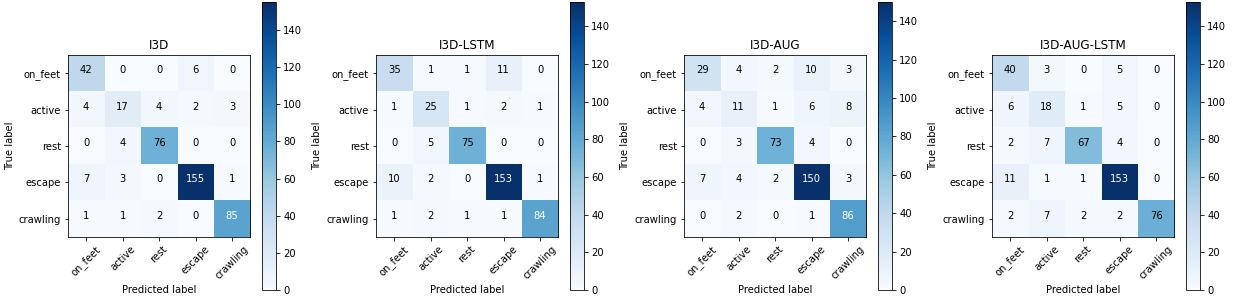}
	\caption{The confusion matrices of the trained models for the five seconds video length training.}
	\label{fig:f3}
\end{figure*}

\section{Results and Analysis} \label{sec:resuts}

\begin{figure*}[h]
	\centering
	\includegraphics[width = 10cm, height = 3cm]{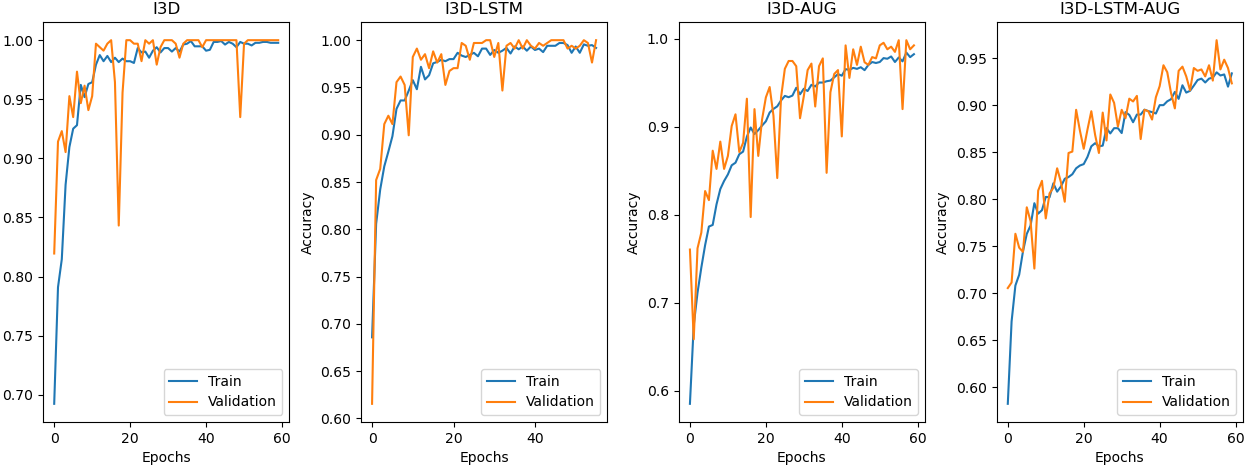}
	\caption{The training versus validation accuracy of the four trained models for the five seconds video length training process. The epochs are represented on the x-axis and the accuracy on the y-axis.}
	\label{fig:f2}
\end{figure*}

In this section, we experimented with benchmark Action Recognition model architectures by comparing the performance of our developed model with the curated dataset for smart baby care.
We modified the I3D architecture by introducing a Conv2D LSTM cell to capture Spatiotemporal features in our curated dataset mapped with the extracted features from the Kinetics 700 human action dataset for our specific Action Recognition task.
We developed four different models to evaluate and compare the performance of each model architecture. These four models are:
\begin{enumerate}
	\item \textbf{Model I}-\textit{Transfer Learning I3D:} This model architecture is based on the popular I3D model with modifications to the Mixed 5c logit section. We removed the classification layer and passed it into flatten and dense layers to train on our new dataset. The model training and testing accuracies are shown in (Table~\ref{Results}), the total number of training parameters is 9,440,261 with a total train time of eleven hours. The RGB kinetics weights were initialized during the training.
	
	\item \textbf{Model II}- \textit{I3D-ConvLSTM2D:} This model architecture added ConvLSTM2D layers to the architecture specified in Model I. The ConvLSTM2D layers were added to extract features on input data combined with time-series information extracted from LSTM cells as described in \cite{donahue2015long}.
	Convolutions are typically used for capturing spatial features in images, and LSTM detects correlations in images over time. Introducing the convolution and LSTM layers were able to capture spatiotemporal features over time. Xingjian et al. \cite{xingjian2015convolutional} proposed the use of convolutional LSTM to achieve better performance by extending fully connected LSTM (FC-LSTM) to include convolutional in the input and state-to-state transitions. The model training and testing accuracies are shown in (Table~\ref{Results}), the total number of training parameters is 9,506,309 with a total train time of ten hours.
	
	\item \textbf{Model III}- \textit{I3D-Augmented Video:}
	The third model architecture has a preprocessing pipeline that performs video augmentation of each dataset with a 50\% probability of flipping, rotating, cropping, adding noise, or occluding portions of the image to expose the model to a more difficult dataset. This approach was implemented to improve the model generalizability and deployment in a real-world scenario where a fine-grained dataset may not be passed into the model. Augmentation strategies on images have been proven to be effective in reducing overfitting and improving model performance. Only a few research works have explored video augmentation for action recognition. We performed video augmentation on our dataset in line with the research work in \cite{yun2020videomix, nida2022video,tanaka2020video}. 
	The model training and testing accuracies are shown in (Table~\ref{Results}), the total number of training parameters is 2,360,069 with a total train time of one day and seventeen hours.
	
	\item \textbf{Model IV}- \textit{I3D-Conv2DLSTM Augmented Video:}
	This model is based on the architecture described in model II with a combination of the video augmentation pre-processing pipeline. We examined the impact of video augmentation on the ConvLSTM2D layer proposed by \cite{xingjian2015convolutional}. The model training and testing accuracies are shown in (Table~\ref{Results}), the total number of training parameters is 9,472,005 with a total train time of one day and ten hours.
\end{enumerate}

We evaluated our model architecture by comparing the model performance with our curated test dataset. For our empirical evaluation of the performance of each of the developed models, we used the same test data. The model with I3D-ConvLSTM2D outperformed all other model architectures. 
As shown in Table [\ref{Results}]. Our first model (I3D-architecture) was trained for eleven hours and had 99\% train accuracy. On the test data, the model accuracy is 90\% with 85\% recall.
The second model architecture with a ConvLSTM2D trained for approximately ten hours with train accuracy of 99\%, the accuracy of 90\% on test data, and recall of 87\%. 
Model I and Model II have relatively similar performance on our Specific AR task. However, Model II with the ConvLSTM2D layer has a slightly better performance and less training time on the dataset. This is further established in the result of the confusion matrix in Figure [\ref{fig:f3}].

The performance of model three after training for one day, seventeen hours on sixty epochs, was 97\% on train data and 84\% on test data with a recall of 75\%.
Model IV trained for about one day, ten hours with a training accuracy of 93\% and accuracy of 85\% on test data with a recall of 81\%.
The lower test accuracy on augmented video is expected as the model was trained on a more complex untrimmed dataset similar to a real-world scenario. Due to class imbalances, the models with augmented video have less performance on test data as shown in our confusion matrix. The model was able to correctly classify raw datasets captured from a live camera. 
Our results showcased a better performance in Action Recognition for smart baby care compared to other benchmark models. The experimental framework developed can be easily deployed for modeling in other specific AR tasks with fewer resources and computation costs.

\section{Limitation}
The proposed work database has been collected and created under grant number 013052 Exhbit A, Procter \& Gamble Company. Therefore, the dataset is available upon request for academic and educational purposes after approval from the funding organization.

\section{Conclusion}
This paper developed a smart home baby monitoring system using a specific task (Smart Baby Care) Action Recognition model architecture. We extracted features from an inflated 3D model based on the Kinetics 400 human actions dataset. The developed model combined spatiotemporal features from the I3D model with 2D convolution LSTM to compute the final feature representation on our curated dataset. The model developed reduced time to production of specific action recognition (SAR) model and computation cost using the pre-trained model to adapt the smart baby care model to a specific task deployment on a new dataset. Additionally, developing dataset for human-specific action is a strenuous activity as most AI models require large dataset to generalize on unseen data. Our model architecture can train on relatively smaller dataset leveraging existing model to extract features using smaller dataset.
\section*{Acknowledgment}
A part of this work has been funded by grant number 013052 Exhbit A, Procter \& Gamble Company, Video Processing project. The authors would like to thank the P\&G team, Oya Aran and Amir Tavanaie for their recommendations and subjective discussions during the project.
\bibliographystyle{ieeetr}
\bibliography{sample-base}

\begin{thebibliography}{10}

\bibitem{Kunze2017}
J.~Kunze, L.~Kirsch, I.~Kurenkov, A.~Krug, J.~Johannsmeier, and S.~Stober,
  ``{Transfer Learning for Speech Recognition on a Budget},'' pp.~168--177,
  Association for Computational Linguistics (ACL), jul 2017.

\bibitem{Fortun2015}
D.~Fortun, P.~Bouthemy, and C.~Kervrann, ``{Optical flow modeling and
  computation: A survey},'' {\em Computer Vision and Image Understanding},
  vol.~134, pp.~1--21, 2015.

\bibitem{Li2020}
A.~Li, M.~Thotakuri, D.~A. Ross, J.~Carreira, A.~Vostrikov, and A.~Zisserman,
  ``{The AVA-Kinetics Localized Human Actions Video Dataset},'' may 2020.

\bibitem{Wang2017}
L.~Wang, Y.~Xiong, Z.~Wang, Y.~Qiao, D.~Lin, X.~Tang, and L.~{Van Gool},
  ``{Temporal Segment Networks for Action Recognition in Videos},'' {\em IEEE
  Transactions on Pattern Analysis and Machine Intelligence}, vol.~41,
  pp.~2740--2755, may 2017.

\bibitem{AccidentDetection}
V.~Adewopo, N.~Elsayed, Z.~ElSayed, M.~Ozer, A.~Abdelgawad, and M.~Bayoumi,
  ``Review on action recognition for accident detection in smart city
  transportation systems,'' 2022.

\bibitem{CDC2021}
CDC, ``{Data and Statistics for SIDS and SUID},'' 2021.

\bibitem{Newell}
A.~Newell, K.~Yang, and J.~Deng, ``Stacked hourglass networks for human pose
  estimation,'' in {\em Computer Vision -- ECCV 2016} (B.~Leibe, J.~Matas,
  N.~Sebe, and M.~Welling, eds.), (Cham), pp.~483--499, Springer International
  Publishing, 2016.

\bibitem{Soomro2014}
K.~Soomro and A.~R. Zamir, ``{Action recognition in realistic sports videos},''
  {\em Advances in Computer Vision and Pattern Recognition}, vol.~71,
  pp.~181--208, 2014.

\bibitem{lan2011discriminative}
T.~Lan, Y.~Wang, and G.~Mori, ``Discriminative figure-centric models for joint
  action localization and recognition,'' in {\em 2011 International conference
  on computer vision}, pp.~2003--2010, IEEE, 2011.

\bibitem{Zhou}
B.~Zhou, A.~Andonian, A.~Oliva, and A.~Torralba, ``Temporal relational
  reasoning in videos,'' in {\em Proceedings of the European conference on
  computer vision (ECCV)}, pp.~803--818, 2018.

\bibitem{pirsiavash2014parsing}
H.~Pirsiavash and D.~Ramanan, ``Parsing videos of actions with segmental
  grammars,'' in {\em Proceedings of the IEEE conference on computer vision and
  pattern recognition}, pp.~612--619, 2014.

\bibitem{gaidon2014activity}
A.~Gaidon, Z.~Harchaoui, and C.~Schmid, ``Activity representation with motion
  hierarchies,'' {\em International journal of computer vision}, vol.~107,
  no.~3, pp.~219--238, 2014.

\bibitem{wang2015transfer}
D.~Wang and T.~F. Zheng, ``Transfer learning for speech and language
  processing,'' in {\em 2015 Asia-Pacific Signal and Information Processing
  Association Annual Summit and Conference (APSIPA)}, pp.~1225--1237, IEEE,
  2015.

\bibitem{chen2015}
D.~Chen and B.~Mak, ``Multitask learning of deep neural networks for
  low-resource speech recognition,'' {\em IEEE/ACM Transactions on Audio,
  Speech, and Language Processing}, vol.~23, pp.~1--1, 07 2015.

\bibitem{Heigold}
G.~Heigold, V.~Vanhoucke, A.~Senior, P.~Nguyen, M.~Ranzato, M.~Devin, and
  J.~Dean, ``Multilingual acoustic models using distributed deep neural
  networks,'' in {\em 2013 IEEE International Conference on Acoustics, Speech
  and Signal Processing}, pp.~8619--8623, May 2013.

\bibitem{weiss2016survey}
K.~Weiss, T.~M. Khoshgoftaar, and D.~Wang, ``A survey of transfer learning,''
  {\em Journal of Big data}, vol.~3, no.~1, pp.~1--40, 2016.

\bibitem{harel2010learning}
M.~Harel and S.~Mannor, ``Learning from multiple outlooks,'' {\em arXiv
  preprint arXiv:1005.0027}, 2010.

\bibitem{daume2009frustratingly}
H.~Daum{\'e}~III, ``Frustratingly easy domain adaptation,'' {\em arXiv preprint
  arXiv:0907.1815}, 2009.

\bibitem{Lim2018}
Y.~K. Lim, Z.~Liao, S.~Petridis, and M.~Pantic, ``Transfer learning for action
  unit recognition,'' {\em arXiv preprint arXiv:1807.07556}, 2018.

\bibitem{Ji2013}
S.~Ji, W.~Xu, M.~Yang, and K.~Yu, ``{3D Convolutional neural networks for human
  action recognition},'' {\em IEEE Transactions on Pattern Analysis and Machine
  Intelligence}, vol.~35, no.~1, pp.~221--231, 2013.

\bibitem{gohil20135g}
A.~Gohil, H.~Modi, and S.~K. Patel, ``5g technology of mobile communication: A
  survey,'' in {\em 2013 international conference on intelligent systems and
  signal processing (ISSP)}, pp.~288--292, IEEE, 2013.

\bibitem{haring2021review}
O.~Haring, S.~W. Azumah, and N.~Elsayed, ``A review of network evolution
  towards a smart connected world,'' {\em arXiv preprint arXiv:2105.13964},
  2021.

\bibitem{elsayed2020reduced}
N.~Elsayed, A.~S. Maida, and M.~Bayoumi, ``Reduced-gate convolutional long
  short-term memory using predictive coding for spatiotemporal prediction,''
  {\em Computational Intelligence}, vol.~36, no.~3, pp.~910--939, 2020.

\bibitem{lim2016speech}
W.~Lim, D.~Jang, and T.~Lee, ``Speech emotion recognition using convolutional
  and recurrent neural networks,'' in {\em 2016 Asia-Pacific signal and
  information processing association annual summit and conference (APSIPA)},
  pp.~1--4, IEEE, 2016.

\bibitem{adewopo2020exploring}
V.~Adewopo, B.~Gonen, and F.~Adewopo, ``Exploring open source information for
  cyber threat intelligence,'' in {\em 2020 IEEE International Conference on
  Big Data (Big Data)}, pp.~2232--2241, IEEE, 2020.

\bibitem{elsayed2018deep}
N.~Elsayed, A.~S. Maida, and M.~Bayoumi, ``Deep gated recurrent and
  convolutional network hybrid model for univariate time series
  classification,'' {\em arXiv preprint arXiv:1812.07683}, 2018.

\bibitem{2021exploring}
V.~A. Adewopo, {\em Exploring Open Source Intelligence for cyber threat
  Prediction}.
\newblock PhD thesis, University of Cincinnati, 2021.

\bibitem{Robles-Serrano2021}
S.~Robles-Serrano, G.~Sanchez-Torres, and J.~Branch-Bedoya, ``{Automatic
  detection of traffic accidents from video using deep learning techniques},''
  {\em Computers}, vol.~10, nov 2021.

\bibitem{shi2015convolutional}
X.~Shi, Z.~Chen, H.~Wang, D.-Y. Yeung, W.-K. Wong, and W.-c. Woo,
  ``Convolutional lstm network: A machine learning approach for precipitation
  nowcasting,'' {\em Advances in neural information processing systems},
  vol.~28, 2015.

\bibitem{Carreira2017}
J.~Carreira and A.~Zisserman, ``{Quo Vadis, action recognition? A new model and
  the kinetics dataset},'' in {\em Proceedings - 30th IEEE Conference on
  Computer Vision and Pattern Recognition, CVPR 2017}, vol.~2017-Janua,
  pp.~4724--4733, may 2017.

\bibitem{labelbox}
Labelbox, ``{Labelbox Academic Program}.'' \url{https://labelbox.com/academic},
  2020-01-25.

\bibitem{zebari2020comprehensive}
R.~Zebari, A.~Abdulazeez, D.~Zeebaree, D.~Zebari, and J.~Saeed, ``A
  comprehensive review of dimensionality reduction techniques for feature
  selection and feature extraction,'' {\em Journal of Applied Science and
  Technology Trends}, vol.~1, no.~2, pp.~56--70, 2020.

\bibitem{ABHANG201697}
P.~A. Abhang, B.~W. Gawali, and S.~C. Mehrotra, ``Chapter 5 - emotion
  recognition,'' in {\em Introduction to EEG- and Speech-Based Emotion
  Recognition} (P.~A. Abhang, B.~W. Gawali, and S.~C. Mehrotra, eds.),
  pp.~97--112, Academic Press, 2016.

\bibitem{Orenstein}
E.~C. Orenstein and O.~Beijbom, ``Transfer learning and deep feature extraction
  for planktonic image data sets,'' in {\em 2017 IEEE Winter Conference on
  Applications of Computer Vision (WACV)}, pp.~1082--1088, 2017.

\bibitem{shorten2019survey}
C.~Shorten and T.~M. Khoshgoftaar, ``A survey on image data augmentation for
  deep learning,'' {\em Journal of Big Data}, vol.~6, no.~1, pp.~1--48, 2019.

\bibitem{Kopuklu}
O.~K{\"{o}}p{\"{u}}kl{\"{u}}, ``{vidaug: Docs, Tutorials, Reviews |
  Openbase}.'' \url{https://openbase.com/python/vidaug}, 2021-10-14.

\bibitem{domenech2020sing}
A.~Dom{\`e}nech~L{\'o}pez, ``Sing language recognition-asl recognition with
  mediapipe and recurrent neural networks,'' {B.S.} thesis, Universitat
  Polit{\`e}cnica de Catalunya, 2020.

\bibitem{donahue2015long}
J.~Donahue, L.~Anne~Hendricks, S.~Guadarrama, M.~Rohrbach, S.~Venugopalan,
  K.~Saenko, and T.~Darrell, ``Long-term recurrent convolutional networks for
  visual recognition and description,'' in {\em Proceedings of the IEEE
  conference on computer vision and pattern recognition}, pp.~2625--2634, 2015.

\bibitem{xingjian2015convolutional}
S.~Xingjian, Z.~Chen, H.~Wang, D.-Y. Yeung, W.-K. Wong, and W.-c. Woo,
  ``Convolutional lstm network: A machine learning approach for precipitation
  nowcasting,'' in {\em Advances in neural information processing systems},
  pp.~802--810, 2015.

\bibitem{yun2020videomix}
S.~Yun, S.~J. Oh, B.~Heo, D.~Han, and J.~Kim, ``Videomix: Rethinking data
  augmentation for video classification,'' {\em arXiv preprint
  arXiv:2012.03457}, 2020.

\bibitem{nida2022video}
N.~Nida, M.~H. Yousaf, A.~Irtaza, and S.~A. Velastin, ``Video augmentation
  technique for human action recognition using genetic algorithm,'' {\em ETRI
  Journal}, 2022.

\bibitem{tanaka2020video}
K.~Tanaka, ``Video augmentation method for the facilitation of skill learning
  in karate,'' in {\em 2020 IEEE International Conference on Teaching,
  Assessment, and Learning for Engineering (TALE)}, pp.~674--677, IEEE, 2020.

\end{thebibliography}
\end{document}